\documentclass[journal]{IEEEtran}
\usepackage{cite}
\usepackage[pdftex]{graphicx}
\graphicspath{ {./figs/} }

\title{Using an Ancillary Neural Network to Capture Weekends and Holidays in an Adjoint Neural Network Architecture for Intelligent Building Management}

\author{Zhicheng~Ding,
        Mehmet~Kerem~Turkcan,
        and~Albert~Boulanger
}

\newcommand{\eg}{\textit{e}.\textit{g}. }

\begin{document}
\maketitle

\begin{abstract}
The US EIA estimated in 2017 about 39\% of total U.S. energy consumption was by the residential and commercial sectors. Therefore, Intelligent Building Management (IBM) solutions that minimize consumption while maintaining tenant comfort are an important component in 
reducing energy consumption.
A forecasting capability for accurate prediction of indoor temperatures in a planning horizon of 24 hours is essential to IBM. It should predict the indoor temperature in both short-term (\eg 15 minutes) and long-term (\eg 24 hours) periods accurately including weekends, major holidays, and minor holidays. Other requirements include the ability to predict the maximum and the minimum indoor temperatures precisely and provide the confidence for each prediction. To achieve these requirements, we propose a novel adjoint neural network architecture for time series prediction that uses an ancillary neural network to capture weekend and holiday information. We studied four long short-term memory (LSTM) based time series prediction networks within this architecture. We observed that the ancillary neural network helps to improve the prediction accuracy, the maximum and the minimum temperature prediction and model reliability for all networks tested. 
\end{abstract}


%

\section{Introduction}
%
%
%

\begin{figure*}[ht]
  \includegraphics[width=\linewidth]{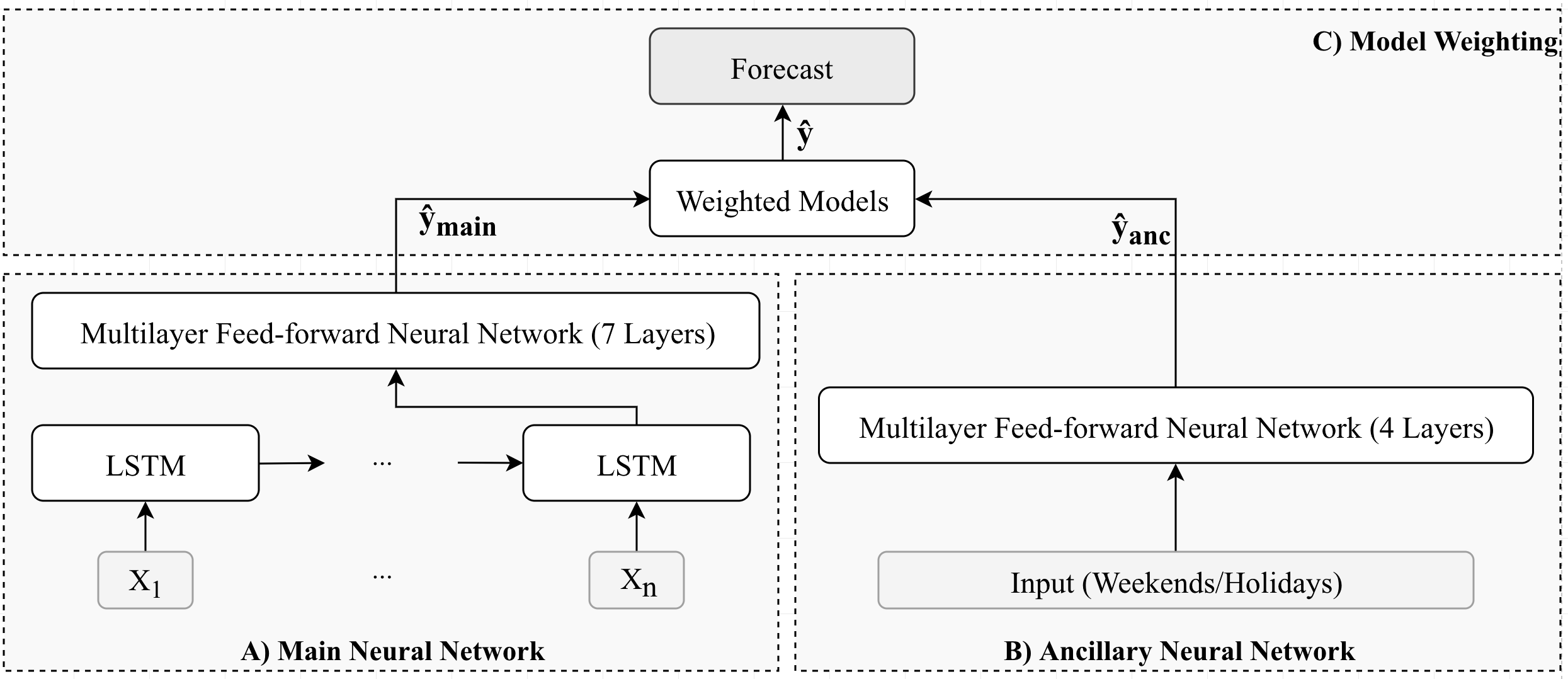}
  \caption{The architecture of the proposed neural network that includes three subnetworks. A) is a main neural network whose inputs are critical features and its outputs are the prediction of each input. It includes an LSTM and a multilayer feed-forward neural network. B) is an ancillary neural network in which extra features are utilized. This module includes a relatively shallow neural network. C) is the model weighting part. We use the weighted average of $\hat{y}_{main}$ and $\hat{y}_{anc}$ to forecast the output. The output includes the predictions of up to 96 steps (24 hours) in the future. In our setting, we use a 2-layer LSTM unit, a 7-layer multilayer feed-forward neural network, and a 4-layer shallow neural network. In total, this model contains 9,349,056 trainable parameters.}
\end{figure*}

\IEEEPARstart{W}{ith} a rapid population growth, the total energy consumption in both residential and commercial buildings has increased and threatens the environment of the earth~\cite{Perez-Lombard2008}. 
Developing an energy-saving IBM system helps to mitigate this problem.

Predicting indoor temperature is the key to building such systems. This is a difficult task because energy consumption is influenced by many different factors such as occupancy, outside weather, solar load, and the building's construction~\cite{Romero2011}. As a result, IBM has become a popular research topic~\cite{Wang2017} in recent years.

Recent literature on IBM focuses on models utilizing artificial neural networks (ANN) since such approaches are powerful in modeling nonlinear problems which are hard to model by other machine learning approaches~\cite{Hippert2001}. Recurrent neural networks (RNNs) have shown remarkable performance in predicting when trained with large time series datasets~\cite{Mikolov2010}. RNNs work well for indoor temperature forecasting because the indoor temperature has certain time-related patterns that for example include daily, weekly, monthly, seasonally,  and yearly patterns. A number of practical studies prove the efficacy of ANNs for IBM. For example, an ANN model was used to predict the air temperature of buildings and was found to exhibit the best performance ~\cite{Mba2016}. Another ANN model that was built for indoor air temperature forecasting outperformed competing regression methods~\cite{Ashtiani2014}.

However, most of these studies utilize a single ANN model~\cite{Wang2017} and only a few studies that combine two or more networks or models have so far been proposed. In addition, only a small set of studies provide the confidence of the model, which is important for evaluating a model's reliability~\cite{Zhu2017}. In addition, the model should predict the indoor temperature in both short-term and the long-term time periods ~\cite{Marvuglia2014}. Short-term predictions provide relatively precise results to the IBM whereas the long-term predictions offer an overview to the system, allowing the IBM enough foresight to optimize current actions to take care of temperature control challenges later in the day.

Inspired by ensemble methods and dropout inference~\cite{Li2017,Gal2016}, we propose a novel adjoint neural network architecture that uses an ancillary neural network to capture weekends and holidays to increase the indoor temperature prediction accuracy for IBM. This architecture also combines the advantages of LSTMs and multilayer feed-forward neural networks~\cite{Svozil1997}. Our proposed method addresses the aforementioned issues and outperforms the prediction of a single model. Meanwhile, it provides predictions for every 15 minutes of 24 hours with 68\% and 95\% confidence intervals (CIs). The main contributions of this paper are summarized as follows.

\begin{itemize}
  \item We propose an adjoint neural network architecture that uses an ancillary neural network to capture weekends and holidays to increase the accuracy of indoor temperature predictions. This architecture enables the prediction of up to 96 steps (24 hours) and provides confidence for all the predictions.
  \item We conducted a comprehensive analysis and comparison by applying our proposed architecture to four popular time series prediction models. The comparison includes three metrics (average error of prediction, the error of max/min temperature prediction, and reliability of the prediction) in both one-step ahead prediction and multiple-steps ahead prediction.
\end{itemize}


\section{Related Work}

\subsection{Artificial Neural Networks}
ANNs have achieved a great success in many different fields~\cite{Dahl2012}. A typical neural network contains many artificial neurons which are known as units. There are three types of units~\cite{LeCun2015}: input units, hidden units, and output units. With more and more hidden layers employed, the neural work is more able to learn deeper abstractions and learns as an artificial brain~\cite{Wang2017}.

Techniques to effectively train ANNs with more than three layers have become standard. \textbf{Deep Neural Network}s (DNNs) are neural networks with more than one hidden layer and it usually works better than shallow neural networks~\cite{GoodfellowIanBengioYoshuaCourville2016} for a particular task. 

DNNs have been used in many different domains. In the image processing domain, DNNs were used to learn graphics representations~\cite{Cao2016} and estimate human pose~\cite{Toshev2014}. In the traffic domain, DNNs were used to classify traffic signs~\cite{Ciresan2012} and predict traffic flow~\cite{Huang2014}. In our IBM domain, DNNs helped predict indoor temperature~\cite{Romeu2013} and energy consumption~\cite{Kalogirou2000}.

In recent years, other ANN designs have become commonly used. The LSTM and Bayesian Neural Network (BNN) designs are relevant and are reviewed below.

\subsection{Long Short-Term Memory}
LSTM networks~\cite{Hochreiter1997} have become popular in the recent years for time series prediction~\cite{Zhu2017}. LSTM networks are a special kind of RNNs which are designed to solve the long-term memory problem. LSTMs have been used in many time series problems. For example, networks using LSTMs have been successfully employed in a number of important problems like speech recognition~\cite{Chan2016}, machine translation~\cite{Cui2016}, and energy load forecasting~\cite{Marino2016}.

In this paper, we will use a densely connected multilayer feed-forward network after an LSTM encoder. The multilayer feed-forward neural network utilizes the temporal information for a robust prediction.  We take the advantage of the encoder in the LSTM layer for extracting temporal information. We feed the final states of the LSTM encoder to the multilayer feed-forward network.

\begin{figure*}[ht]
  \includegraphics[width=\linewidth,]{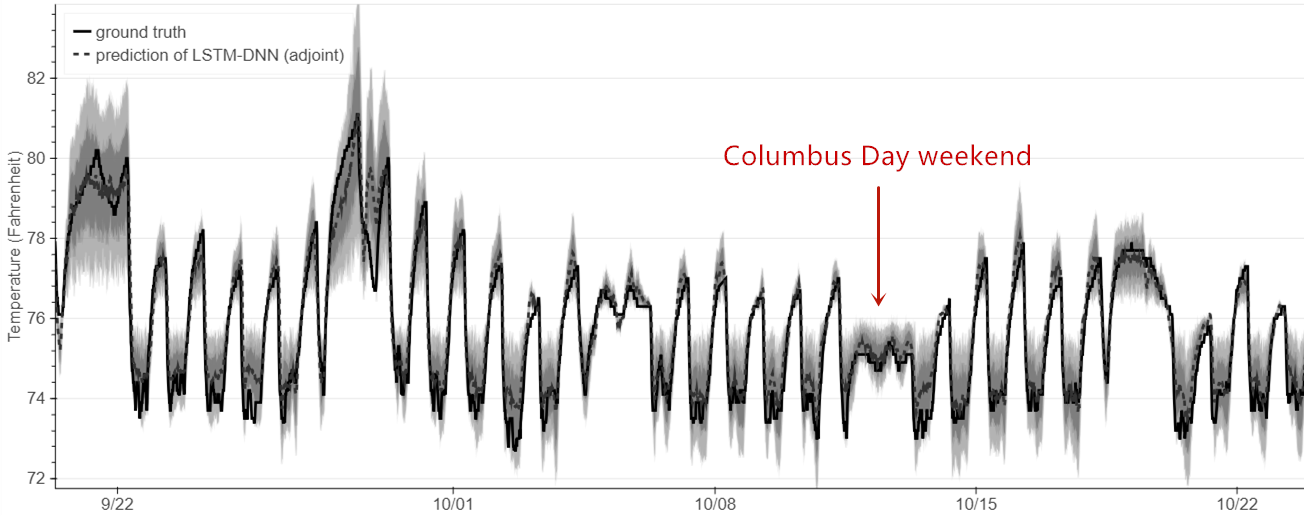}
  \caption{Sample of one-step ahead prediction. The 68\% and 95\% CIs are shown as gray and light gray color bands respectively. The black solid line represents the ground truth indoor temperature. The gray dash line denotes the prediction of LSTM-DNN. The indoor temperature shows in a different pattern as it is on weekdays. As for the data used in our work, Columbus day was the 13th of October.}

\end{figure*}

\begin{figure*}[ht]
  \centering
  \includegraphics[width=\linewidth,]{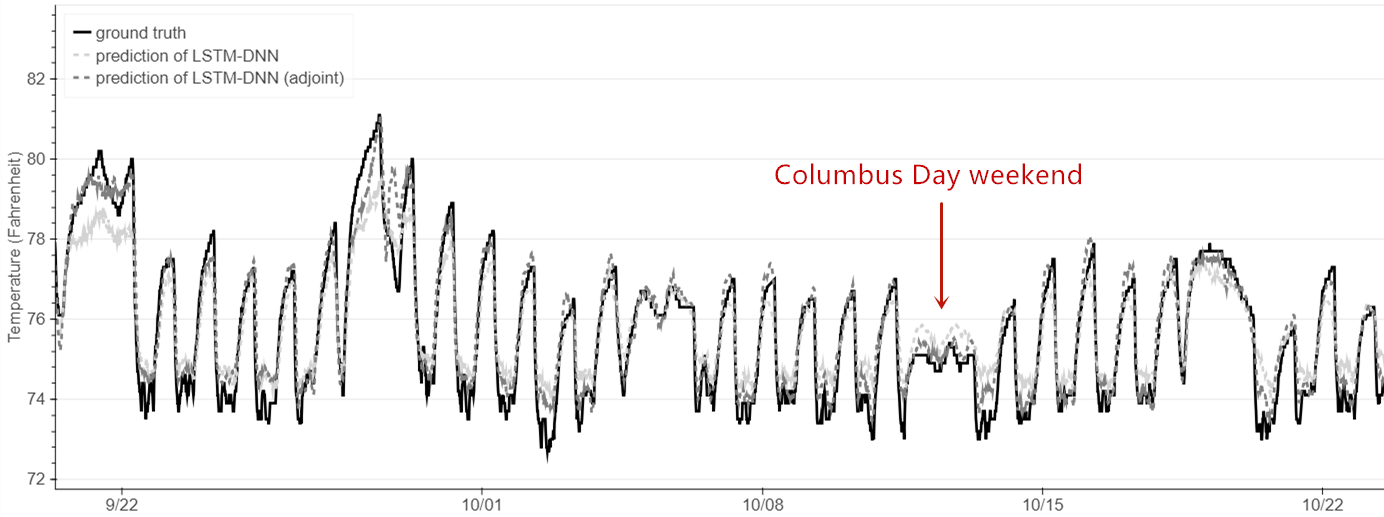}
  \caption{Comparison of one-step ahead prediction by the LSTM-DNN without the ancillary network and the adjoint LSTM-DNN. The light gray dash line represents the prediction by the LSTM-DNN without the ancillary neural network, and the gray dash line denotes the prediction by the adjoint LSTM-DNN. The indoor temperature shows in a different pattern as it is on weekdays. As for the data used in our work, Columbus day was the 13th of October.}

\end{figure*}

\begin{figure}[h]
  \centering
    \includegraphics[width=\linewidth, height=4cm]{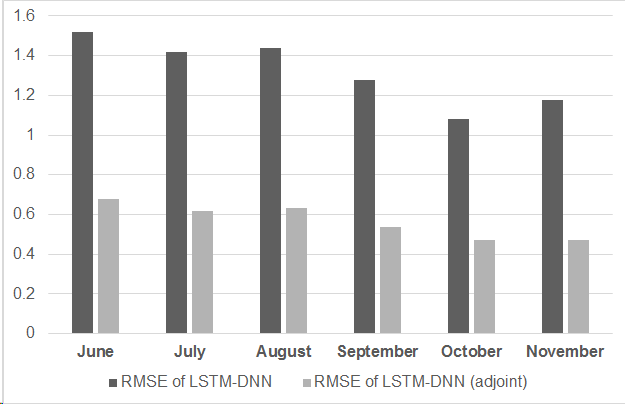} \\(a) RMSE of all predictions
    \includegraphics[width=\linewidth, height=4cm]{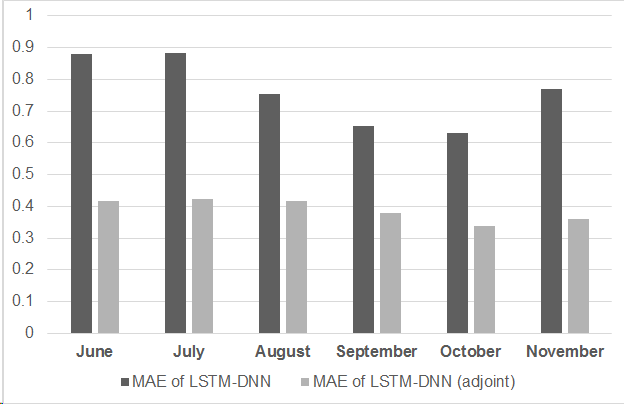} \\(b) MAE of all predictions
    \includegraphics[width=\linewidth, height=4cm]{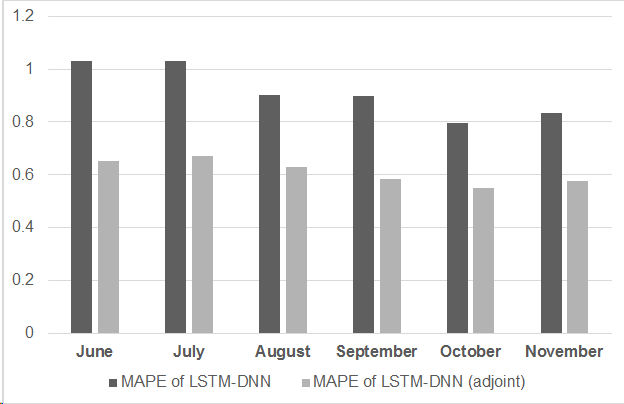} \\(c) MAPE of all predictions
  \caption{One-step ahead prediction's error of RMSE (a), MAE (b), and MAPE (c) of all predictions.}
  \end{figure} 
  
\begin{figure}[h]
  \centering
    \includegraphics[width=\linewidth, height=4cm]{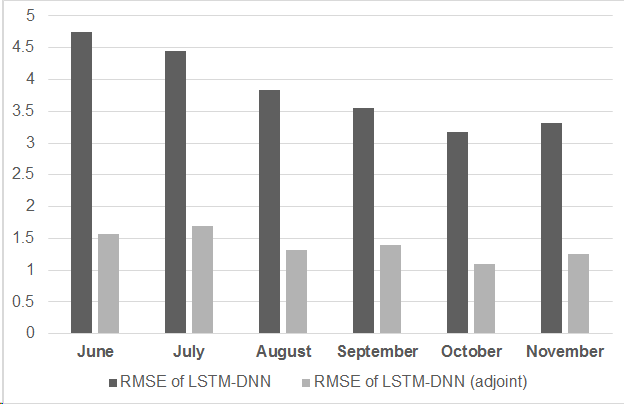} \\(a) RMSE of max/min temperature predictions
    \includegraphics[width=\linewidth, height=4cm]{one_step_ahead_max_min_temp_RMSE.png} \\(b) MAE of max/min temperature predictions
    \includegraphics[width=\linewidth, height=4cm]{one_step_ahead_max_min_temp_RMSE.png} \\(c) MAPE of max/min temperature predictions
  \caption{One-step ahead prediction's  error of RMSE (a), MAE (b), and MAPE (c) of max/min temperature predictions.}
  \end{figure}

  \begin{figure}[h]
    \centering
    \begin{minipage}{.47\textwidth}
      \centering
    \includegraphics[width=\linewidth,]{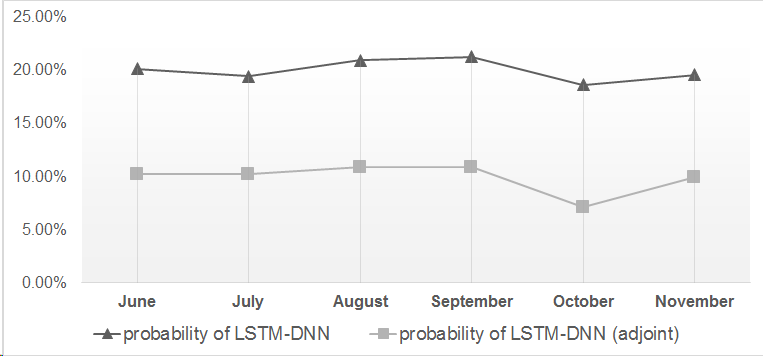} \\ (a) Probability of predictions are not within 68\% CI
    \end{minipage}
    \begin{minipage}{.47\textwidth}
      \centering
    \includegraphics[width=\linewidth,]{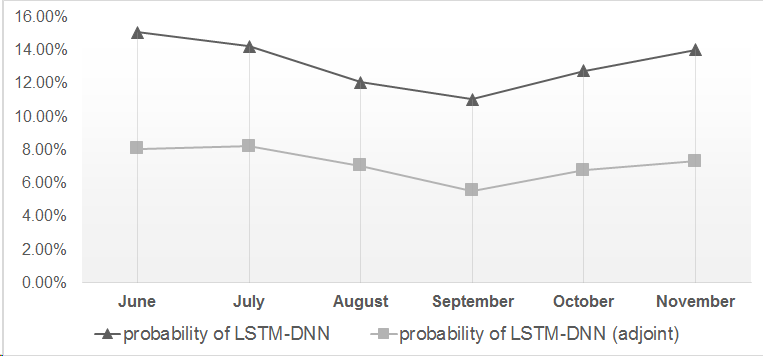}\\ (b) Probability of predictions are not within 95\% CI
    \end{minipage}
    \caption{The monthly probability of one-step ahead predictions that are not within 68\% CI (a) and 95\% CI (b).}
  \end{figure}

  \begin{figure}[h]
    \centering
    \includegraphics[width=\linewidth]{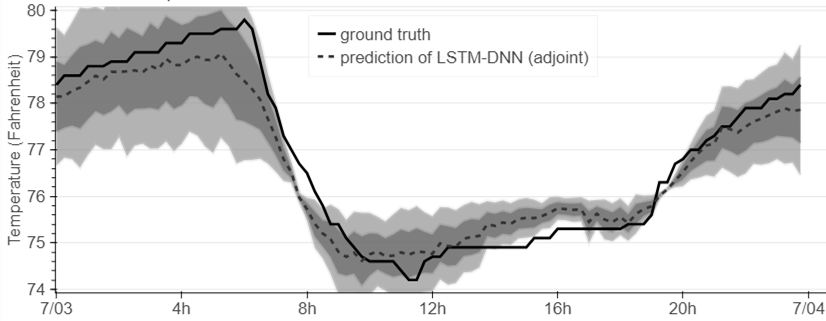}
    \caption{Sample prediction of 96-steps ahead (24 hours) scenario. The 68\% and 95\% CIs are shown as gray and light gray color bands respectively. The black solid line denotes the ground truth indoor temperature and gray dash line represents prediction by adjoint model.}
  \end{figure}

\begin{figure*} [h]
  \centering
  \begin{minipage}{.49 \textwidth}
    \centering
    \includegraphics[width=\linewidth]{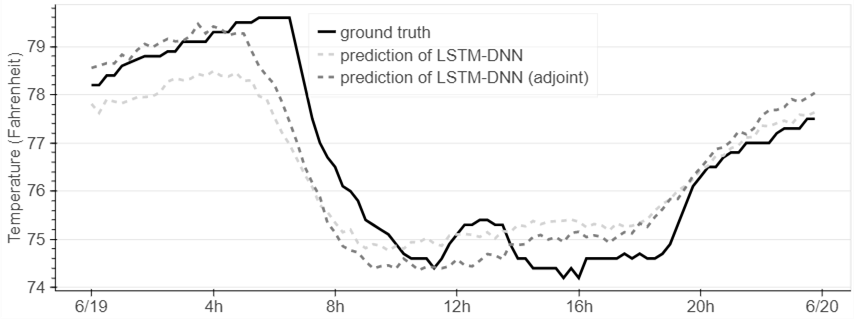} \\ (a) 96-steps ahead prediction in 19th June
  \end{minipage}%
  \begin{minipage}{.49\textwidth}
    \centering
    \includegraphics[width=\linewidth]{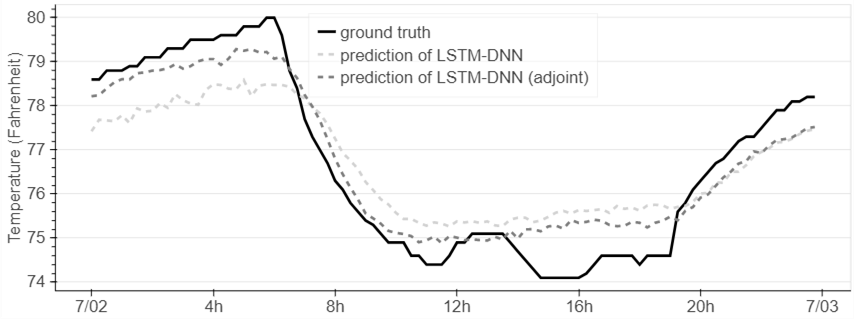} \\ (b) 96-steps ahead prediction in 2nd July
  \end{minipage}%
  
  \begin{minipage}{.49\textwidth}
    \centering
    \includegraphics[width=\linewidth]{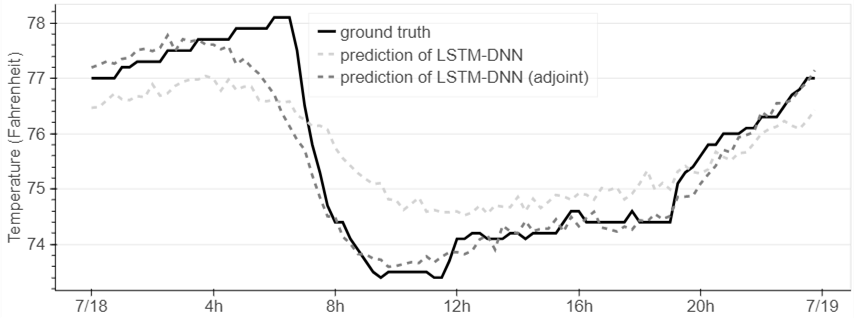} \\ (c) 96-steps ahead prediction in 17th July
  \end{minipage}
  \begin{minipage}{.49\textwidth}
    \centering
    \includegraphics[width=\linewidth]{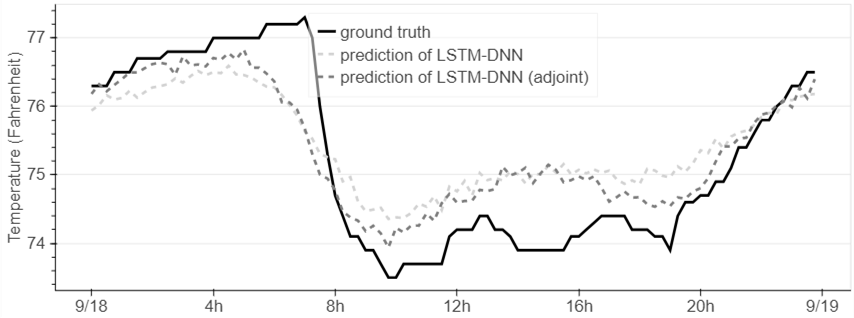} \\ (d) 96-steps ahead prediction in 18th September
  \end{minipage}
  \caption{Comparison of multi-step ahead prediction results between the LSTM-DNN without the ancillary network and the adjoint LSTM-DNN. The light gray dash line represents the prediction by the LSTM-DNN without the ancillary neural network, and the gray dash line denotes the predictions by the adjoint LSTM-DNN.}
  \end{figure*}

\section{Methodology}
The complete architecture proposed is shown in Figure 1. The architecture contains three parts: A) is the main neural network (LSTM followed by a multilayer feed-forward network) which aims to learn temporal patterns from significant features. B) is an ancillary neural network which takes in extra features regarding weekends and holidays using an LSTM followed by relatively shallow multilayer feed-forward network). C) is the combiner part whose outputs are the weighted average of the predictions from part (A) and part (B). The output includes the forecast of one-step ahead (next 15 minutes) and multi-step ahead (up to 96 steps, 24 hours in total).  After that, we add a dropout unit to infer the CIs of the result. We will discuss each module and CI in the details below.

\subsection{Main Neural Network}
This module aims to learn from the important features. Those features usually change along with time, such as outside temperature, occupancy, and date. Also, these features have daily, weekly, monthly, seasonally, and yearly patterns. Thus, we construct the data as continuous time series input. Those data pass to the LSTM units which learn temporal information and construct internal states. The internal states are further propagated to a 7-layer feed-forward network that forecasts the one-step ahead and 96-steps ahead indoor temperature $\hat{y}_{main}$.

Given a dataset which includes $S$ timestamps, the model needs to predict the next $n$ timestamps and for each timestamp, there are $m$ features. Thus, we construct the input data as a three-dimensional matrix and the size of the matrix is $S \times n \times m$, 

We use a 2-layer LSTM network to extract temporal information and then feed the temporal information to a 7-layer feed-forward neural network. Next, the network forecasts the indoor temperature $\hat{y}_{main}$ which contains 96-steps ahead predictions. Since the duration of each time step is 15 minutes, the 96-steps ahead predictions will cover the prediction of every 15 minutes of next 24 hours.

\subsection{Ancillary Neural Network}
In the ancillary neural network, we aim to utilize extra features like weekends and holidays. These features are ancillary because, with enough data, this information could be learned by the main neural network itself from significant features fed into the main neural network. But explicitly providing these features helps increase the performance of the model especially if the size of the dataset is small. These values of these extra features are either 0 or 1 which is used as an indicator of being a weekend or holiday. Then, to reduce the model complexity, those extra features are learned by a relatively shallow neural network (a 4-layer feed-forward neural network). The size and dimension of the output $\hat{y}_{anc}$ is exactly the same as the output from the main neural network.

\subsection{Model Weighting}
After the main neural network and ancillary neural network are fully trained, we use the weighted average~\cite{Qiu2014} and use \textbf{rectifier} (ReLU)~\cite{Glorot2011} as the activation function to forecast the output, shown as below:

\begin{equation}
\hat{y} = ReLU(w_1 \hat{y}_{main} + w_2 \hat{y}_{anc})
\end{equation}

\noindent where $ReLU$ is the activation function which is computationally efficient to compute and has less likelihood of a vanishing gradient.

The output of the weighted average $\hat{y}$ includes multiple steps predictions with the same dimension as the output of main neural network module $\hat{y}_{main}$ and ancillary neural network $\hat{y}_{anc}$. We minimize the root mean squared error (RMSE) between all these values and corresponding ground truth value. In this case, we had prepared our ground truth indoor temperature with multiple timestamps as a three-dimensional matrix.

The output layer provides one output for each of the 96 timestamps. The closer timestamp to the forecast time is usually more accurate than later timestamps. The later timestamp forecasting provides an overview of how indoor temperature is going to change and allows for taking some actions ahead of time~\cite{Chang2007} for better anticipatory HVAC control of the building. This helps provide desired temperatures with less energy consumed since less radical (more planned) actions are taken.

\subsection{Deriving Confidence Intervals}
After the model is fully trained, we need to derive the CI of the output of the model. We use MC dropout proposed in~\cite{Gal2015} and this provides a framework to estimate uncertainty without any change of the existing model.

Specifically, we add stochastic dropouts to each hidden layer of the neural network. Then we add sample variance to the output from the model~\cite{Li2017}. Finally, we estimate the uncertainty by approximating the sample variance. We assume that indoor temperature is approximately a Gaussian distribution. In this paper, we will derive the 68\% and 95\% CIs and use it to evaluate the reliability of the model. Given the sample data, we could calculate the mean $\mu$ and standard deviation $\sigma$. Then the $100 \cdot (1-\alpha)\%$ CI~\cite{Penciana2004} is:

\begin{equation}
  \Big[\mu - t_{1-\alpha/2} \cdot \frac{\sigma}{\sqrt{n}}, \mu + t_{1-\alpha/2} \cdot \frac{\sigma}{\sqrt{n}} \Big]
\end{equation}
where $n$ is the sample number and $t$ denotes the density function.

\begin{figure}[h]
  \centering
    \includegraphics[width=\linewidth, height=4cm]{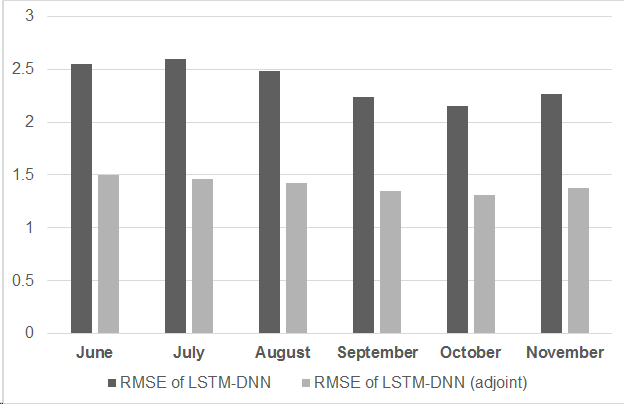} \\(a) RMSE of all predictions
    \includegraphics[width=\linewidth, height=4cm]{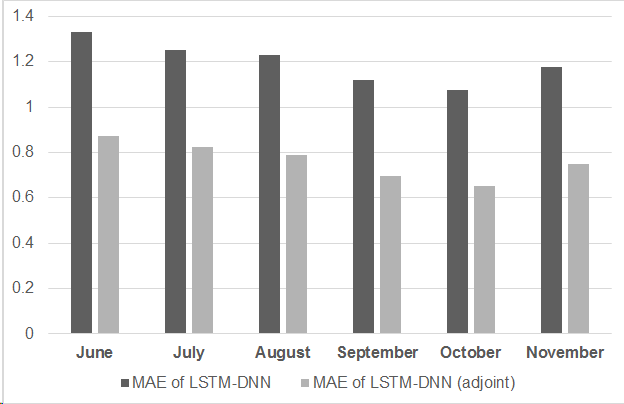} \\(b) MAE of all predictions
    \includegraphics[width=\linewidth, height=4cm]{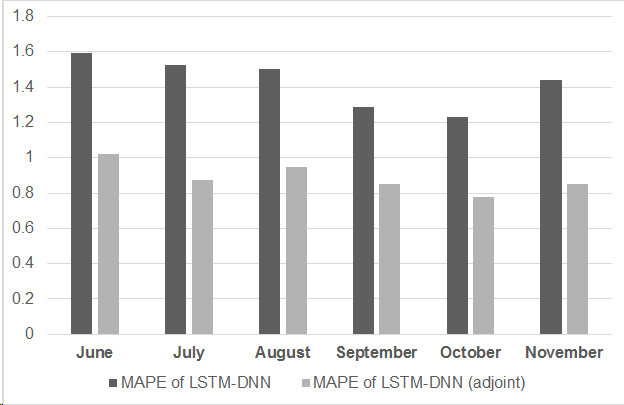} \\(c) MAPE of all predictions
  \caption{Multi-step ahead prediction's error of RMSE (a), MAE (b), and MAPE (c) of all predictions.}
  \end{figure} 
  
\begin{figure}[h]
  \centering
    \includegraphics[width=\linewidth, height=4cm]{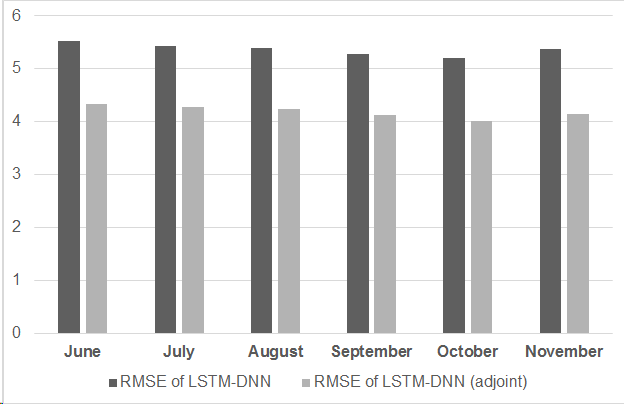} \\(a) RMSE of max/min temperature predictions
    \includegraphics[width=\linewidth, height=4cm]{multi_step_ahead_max_min_temp_RMSE.png} \\(b) MAE of max/min temperature predictions
    \includegraphics[width=\linewidth, height=4cm]{multi_step_ahead_max_min_temp_RMSE.png} \\(c) MAPE of maxi/min temperature predictions
  \caption{Multi-step ahead prediction's  error of RMSE (a), MAE (b), and MAPE (c) of max/min temperature predictions.}
  \end{figure}  

  \begin{figure}[h]
    \centering
    \begin{minipage}{.47\textwidth}
      \centering
    \includegraphics[width=\linewidth,]{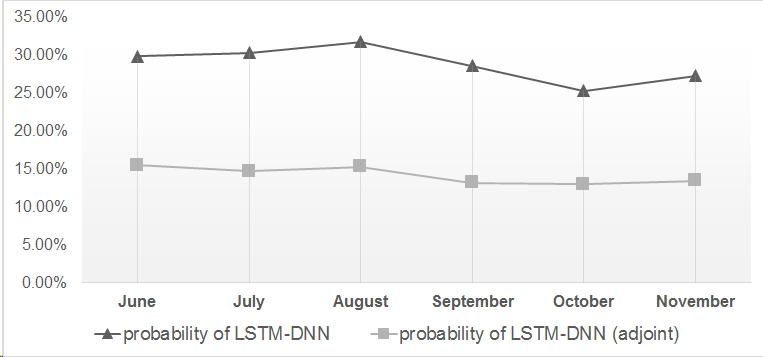} \\ (a) Probability of predictions are not within 68\% CI
    \end{minipage}
    \begin{minipage}{.47\textwidth}
      \centering
    \includegraphics[width=\linewidth,]{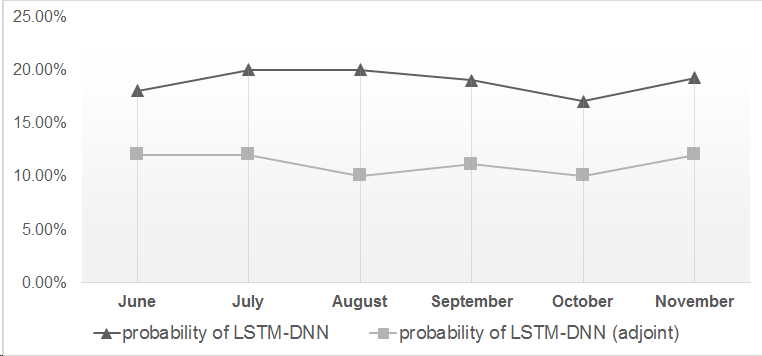}\\ (b) Probability of predictions are not within 95\% CI
    \end{minipage}
    \caption{The monthly probability of multi-step ahead predictions that are not within 68\% CI (a) and 95\% CI (b).}
  \end{figure}

\begin{table*} [h]
  \caption {One-step ahead prediction comparison between the model Using the adjoint architecture and the model without the ancillary network.} 
  \label{tab:title} 
\centering
 \begin{tabular}{||l | c c c | c c c | c c ||} 
    \hline
    \multicolumn{1}{||c|}{Models} &
    \multicolumn{3}{c|}{Error of All Predictions}&
    \multicolumn{3}{c|}{Error of Max/Min Predictions}&
    \multicolumn{2}{c||}{GT not within CI}\\
   & RMSE & MAE & MAPE & RMSE & MAE & MAPE & 68\% & 95\%
 \\ [0.5ex] 
 \hline\hline 
 LSTM-RNN & 2.96 & 1.33 & 1.73 & 4.23 & 1.87 & 1.69 & 24.79\% & 15.43\% \\%
 \hline
 LSTM-RNN (adjoint) & 2.65 & 1.20 & 1.55 & 2.82 & 1.23 & 1.84 & 20.13\% & 11.69\%\\ %
 \hline
 \hline
 LSTM-encoder-decoder & 4.22 & 1.77 & 2.32 & 6.97 & 3.11 & 4.05 & 43.70\% & 19.19\% \\%
 \hline
 LSTM-encoder-decoder (adjoint) & 1.06 & 0.63 & 0.83 & 3.25 & 1.45 & 1.83 & 23.80\% & 13.56\%\\ %
 \hline
 \hline
 LSTM-encoder w/predictnet & 1.12 & 0.66 & 0.86 & 3.58 & 1.54 & 1.95 & 41.75\% & 20.85\% \\%
 \hline
 LSTM-encoder w/predictnet (adjoint) & 0.83 & 0.60 & 0.75 & 2.28 & 1.31 & 1.61 & 19.82\% & 13.82\%\\ %
 \hline
 \hline
 LSTM-DNN & 1.32 & 0.76 & 0.91 & 3.84 & 1.79 & 2.21 & 19.95\% & 13.18\% \\%
 \hline
 \textbf{LSTM-DNN (adjoint)} &  \textbf{0.57} &  \textbf{0.39} &  \textbf{0.61} &  \textbf{1.39} &  \textbf{0.78} &  \textbf{1.26} &  \textbf{9.84\%} &  \textbf{7.15\%} \\ %
 \hline
\end{tabular}
\end{table*}

  \begin{table*} [h]
    \caption {Multi-steps ahead Prediction Comparison between the model Using the adjoint architecture and the model without the ancillary network} 
    \label{tab:title} 
  \centering
   \begin{tabular}{||l | c c c | c c c | c c ||} 
      \hline
      \multicolumn{1}{||c|}{Models} &
      \multicolumn{3}{c|}{Error of All Predictions}&
      \multicolumn{3}{c|}{Error of Max/Min Predictions}&
      \multicolumn{2}{c||}{GT not within CI}\\
     & RMSE & MAE & MAPE & RMSE & MAE & MAPE & 68\% & 95\% 
   \\ [0.5ex] 
   \hline\hline 
   LSTM-RNN & 3.83 & 1.70 & 2.20 & 5.58 & 2.59 & 3.39 & 33.68\% & 25.08\% \\%
   \hline
   LSTM-RNN (adjoint) & 2.93 & 1.39 & 1.85 & 4.38 & 2.09 & 2.31 & 24.05\% & 11.66\%\\ %
   \hline
   \hline
   LSTM-encoder-decoder & 6.21 & 2.23 & 2.91 & 8.03 & 3.41 & 3.77 & 47.71\% & 21.68\% \\%
   \hline
   LSTM-encoder-decoder (adjoint) & 2.02 & 0.99 & 1.29 & 4.98 & 2.25 & 2.92 & 26.31\% & 15.70\%\\ %
   \hline
   \hline
   LSTM-encoder w/predictnet & 2.07 & 1.02 & 1.32 & 4.86 & 2.26 & 2.94 & 47.78\% & 24.22\% \\%
   \hline
   LSTM-encoder w/predictnet (adjoint) & 1.72 & 0.99 & 1.28 & 4.24 & 2.04 & 2.65 & 19.94\% & 15.94\%\\ %
   \hline
   \hline
   LSTM-DNN & 2.38 & 1.19 & 1.43 & 5.37 & 2.40 & 2.92 & 28.79\% & 18.87\% \\%
   \hline
   \textbf{LSTM-DNN (adjoint)} &  \textbf{1.40} &  \textbf{0.76} &  \textbf{0.88} &  \textbf{4.19} &  \textbf{1.88} &  \textbf{2.28} &  \textbf{14.15\%} &  \textbf{11.18\%} \\ %
   \hline
  \end{tabular}
  \end{table*}

\section{Experiments}
In this section, we will first test our proposed architecture with the LSTM-DNN~\cite{Sainath2015} network in both one-step ahead (15 minutes) prediction and 96-steps ahead (24 hours) prediction. Later, we will do similar test to the other three popular time series prediction models: LSTM RNN~\cite{Hochreiter1997}, LSTM encoder-decoder~\cite{Cho2014}, and LSTM encoder w/predictnet~\cite{Zhu2017}. We will compare not only the error (RMSE, MAE, and MAPE) of all predictions but also the error of predicting the maximum and the minimum temperatures. In addition, we will evaluate the reliability of the models by calculating the probability of ground truth temperature being within the 68\% and 95\% CIs respectively.

\subsection{Setting}
The data comes from a multistory building located in New York City (NYC). Data are collected by the Building Management System (BMS) every 15 minutes. The temperature predicted is for one sensor on a floor of the building. Total building occupancy data are also collected. In addition, we collect the outside weather information, including wind speed and direction, humidity and dew point, pressure and weather status (fog, rain, snow, hail, thunder, and tornado),  and temperature from the Central Park NOAA weather station. The analyzed data are from June 9th, 2012 to November 16th, 2014 (84,768 timestamps). We utilized 67,814 of the timestamps for training and 16,954 for testing. The ratio of training to testing is 8:2. We then split 20\% of the timestamps from the training dataset for validation.  After the neural networks are trained, we added dropout layer and repeatedly sample for 10,000 times. Finally, we derive 68\% and 95\% CIs from the sample outputs.

\subsection{One-Step Ahead Prediction Evaluation}

We first compare the capacity of capturing time series pattern between the LSTM-DNN model without the ancillary model and the LSTN-DNN model using the adjoint architecture. Later on, we will evaluate the improvement of using our proposed architecture from three different perspectives (error of the total prediction, error of predicting max/min temperature, and model reliability based on the CIs).

\noindent\textbf{Capacity of Capturing the Time Series Pattern:} 
to retrieve the one-step ahead data from the output consisting of 96-steps ahead predictions, we extract the first step of the output temperature of each input and splice these data. Figure 2 demonstrates the sample predictions of using our proposed architecture. The plot includes one-step ahead prediction with the 68\% CI and 95\% CI. The 68\% and 95\% CIs are shown as gray and light gray color band respectively. The black solid line represents the ground truth indoor temperature and the gray dash line is the prediction. We noticed that the model learns the daily and weekly patterns as the temperature pattern during weekdays and weekends is different.  But we also found that the bands of the CIs are large at the maximum and the minimum temperature. 

In order to evaluate the improvement of prediction using the adjoint neural network, we plot the predictions (without CIs) of the LSTM-DNN without the ancillary network, the adjoint LSTM-DNN, and ground truth indoor temperature in the same plot. Figure 3 shows the selected result of the one-step ahead prediction from 22nd September to 22nd October. We observed that both models capture the daily, weekly and holiday patterns well. But the adjoint model has better predictions, especially when the predicting of the maximum and the minimum temperature. Next, we will compare the model using our proposed architecture with the model without the ancillary network these three different perspectives.

\subsubsection{Error of All Predictions} 
First of all, we calculated the monthly error of all predictions. Specifically, we calculated the monthly error of RMSE, MAE, and MAPE using all timestamps (every 15 minutes) for the months of June to November. Then we compared the error of both the LSTM-DNN without the ancillary network and the adjoint LSTM-DNN. Figure 4 illustrates the comparison of error (RMSE, MAE, and MAPE) of the two models. It is obvious that the model using our proposed architecture has a much smaller error than the LSTM-DNN without the ancillary network.

\subsubsection{Error of Max/Min Predictions}
In addition, we evaluated the improvement of using our proposed architecture to predict the maximum and the minimum temperatures. To begin with, we found the timestamps of both the maximum and the minimum temperatures with respect to each day. Next, we calculated the RMSE, MAE, and MAPE from 30 minutes (2 timestamps) before to 30 minutes after the timestamp of the maximum or the minimum temperatures. Last, we summarized the error in the same month. Figure 5 illustrates that, using our proposed architecture, the model has a much better capacity for predicting the maximum and the minimum temperatures.

\subsubsection{Reliability of Predictions}
Last, we evaluated the reliability of the model.  This is evaluated by calculating the probability that the ground truth indoor temperatures are not within 68\% and 95\% CIs respectively, as shown in Figure 6. We notice that both models have similar changes during the month. But the probability of using our proposed architecture is always lower than the LSTM-DNN without the ancillary network. Therefore, using the adjoint network increases the reliability of the model.

Then, the same analysis was conducted for the three other base models, LSTM RNN, LSTM encoder-decoder, and LSTM encoder w/predictnet, within the adjoint neural network architecture and without the adjoint architecture. 

To sum up, Table 1 illustrates the comparison of one-step ahead predictions between the models with and without the ancillary network. It shows that the adjoint neural network architecture decreases the error and increases the reliability of the models. Also, we observed that our proposed model (LSTM-DNN) which is not the best model without the ancillary network becomes the best model when used within our adjoint neural network architecture.

\subsection{Multiple Timestamps Evaluation}

In this section, we will evaluate the prediction errors for 96-steps ahead. Each step represents 15 minutes, so 96 steps denote the predictions for 24 hours. We will first compare the models' performance using or not using our proposed architecture to capture the daily pattern. Then, we will compare the two models from these three different prospectives.

\noindent\textbf{Capacity of Capturing the Time Series Pattern:} 
Figure 7 shows the predictions of using our proposed architecture for July 1st which was a Tuesday. The prediction is for 96-steps ahead including the 68\% CI and the 95\% CI. The 68\% and 95\% CIs are shown as the gray and light gray color bands respectively. The black solid line represents the ground truth indoor temperature and the gray dash line is the scalar predictions of the adjoint model. We noticed the model learned the daily temperature and predicts the 24 hours fairly well. At midnight, the indoor temperature is high since the HVAC system is shut down at that hour. During the daytime, the HVAC system is operating, and the indoor temperature is lower. As people leave the office, the building operators ramp down and later turn off the HVAC system and the temperature raises again.

In order to evaluate the improvement of the prediction using our proposed architecture, we plot the predictions (without CIs) of the LSTM-DNN without the ancillary network, the adjoint LSTM-DNN, and the ground truth indoor temperature in the same plot, as shown in Figure 8. We observed that the adjoint neural network has better performance for fitting the 24 hour prediction, especially for predicting the maximum and the minimum indoor temperature. Next, we compare the model using our proposed architecture with the model without the ancillary network from three different perspective.

\subsubsection{Error for all predictions} 

Firstly, we calculated the monthly error of all predictions. Specifically, we calculated the monthly error of RMSE, MAE, and MAPE using 96-steps ahead (24 hours) predictions of all the timestamps in that month respectively. Next, we compared the error of both the LSTM-DNN without the ancillary network and the LSTM-DNN using our proposed architecture. Figure 9 demonstrates the comparison of the two models. Though the error is higher than the one-step ahead predictions, the model that uses our proposed architecture still outperforms the model without the ancillary network.

\subsubsection{Error of Max/Min Predictions}
Secondly, we evaluated the predictions on the maximum and the minimum temperature. To begin with, we found the timestamps of both the maximum and the minimum temperature. Then we calculated the RMSE, MAE, and MAPE from 30 minutes before and after the time of the maximum or the minimum temperature. Finally, we summarized the error in the same month. Figure 10 illustrates that the model using our proposed architecture has the better capacity to predict the maximum and the minimum temperatures.

\subsubsection{Reliability of Predictions}
Thirdly, we evaluated the reliability of the model. We calculated the probability of the ground truth indoor temperatures are outside of the 68\% CI and the 95\% CI respectively. Thus, the lower the probability is, the more reliable the model is. Figure 11 shows the result of how the probability changes from June to November. We noticed that the models have fairly similar change during this time period, but the adjoint neural network architecture is more reliable than the LSTM-DNN without the ancillary network. 

Then, the same analysis was conducted for the three other base models, LSTM RNN, LSTM encoder-decoder, and LSTM encoder w/predictnet, within the adjoint neural network architecture and without the adjoint architecture.

To sum up, Table 2 illustrates the comparison of the multi-step ahead predictions between the model using our proposed model and the corresponding model without the ancillary network. It shows that the ancillary neural network can also decrease the error and increase the reliability of the model in multi-step ahead forecast. The interesting finding, that the model (LSTM-DNN) which is not the best model becomes the best model by using our adjoint neural network architecture, also holds true in the multi-step ahead forecast.

\section{Conclusions}

In this paper, we propose a novel adjoint neural network architecture for time series prediction that uses an ancillary neural network to capture weekend and holiday information for IBM. We used the dataset of a multistory building in NYC and compared four different base models (LSTM RNN, LSTM encoder-decoder, LSTM encoder w/predictnet and our proposed LSTM-DNN model) within the adjoint architecture with the corresponding models without the ancillary network. The models' performance was evaluated in both one-step ahead prediction (15 minutes) and 96-steps ahead prediction (24 hours) in three different perspectives. First, we compared the total error of RMSE, MAE, and MAPE of all prediction accounted. Second, we compared the error (RMSE, MAE, and MAPE) of predicting the maximum and the minimum temperatures. Third, we compared the reliability of the models by calculating the probability of the ground truth indoor temperatures are not within CIs.

From the result of the one-step ahead prediction (Figure 3), the models using an ancillary NN successfully capture the daily, the weekly patterns and the holidays. In addition, it performs better prediction result than the models without using ancillary network. Then,  from the result of the multi-step ahead prediction (Figure 7), the models using an ancillary network successfully captures the daily patterns better, especially on predicting the maximum and the minimum temperatures.

Our adjoint neural network architecture with the LSTM-DNN model could be used for building a more dependable IBM system. With accountable predicted indoor temperatures, the system can provide comfortable indoor temperatures with less energy consumed.


%





\ifCLASSOPTIONcaptionsoff
  \newpage
\fi



\bibliographystyle{IEEEtran}
\bibliography{ding}

\begin{thebibliography}{10}
\providecommand{\url}[1]{#1}
\csname url@samestyle\endcsname
\providecommand{\newblock}{\relax}
\providecommand{\bibinfo}[2]{#2}
\providecommand{\BIBentrySTDinterwordspacing}{\spaceskip=0pt\relax}
\providecommand{\BIBentryALTinterwordstretchfactor}{4}
\providecommand{\BIBentryALTinterwordspacing}{\spaceskip=\fontdimen2\font plus
\BIBentryALTinterwordstretchfactor\fontdimen3\font minus
  \fontdimen4\font\relax}
\providecommand{\BIBforeignlanguage}[2]{{%
\expandafter\ifx\csname l@#1\endcsname\relax
\typeout{** WARNING: IEEEtran.bst: No hyphenation pattern has been}%
\typeout{** loaded for the language `#1'. Using the pattern for}%
\typeout{** the default language instead.}%
\else
\language=\csname l@#1\endcsname
\fi
#2}}
\providecommand{\BIBdecl}{\relax}
\BIBdecl

\bibitem{Perez-Lombard2008}
L.~P{\'{e}}rez-Lombard, J.~Ortiz, and C.~Pout, ``{A review on buildings energy
  consumption information},'' \emph{Energy and Buildings}, 2008.

\bibitem{Romero2011}
J.~Romero, J.~Navarro-Esbr{\'{i}}, and J.~Belman-Flores, ``{A simplified
  black-box model oriented to chilled water temperature control in a variable
  speed vapour compression system},'' \emph{Applied Thermal Engineering}, 2011.

\bibitem{Wang2017}
\BIBentryALTinterwordspacing
Z.~Wang and R.~S. Srinivasan, ``{A review of artificial intelligence based
  building energy use prediction: Contrasting the capabilities of single and
  ensemble prediction models},'' \emph{Renewable and Sustainable Energy
  Reviews}, vol.~75, no. September 2015, pp. 796--808, 2017. [Online].
  Available: \url{http://dx.doi.org/10.1016/j.rser.2016.10.079}
\BIBentrySTDinterwordspacing

\bibitem{Hippert2001}
H.~Hippert, C.~Pedreira, and R.~Souza, ``{Neural networks for short-term load
  forecasting: a review and evaluation},'' \emph{IEEE Transactions on Power
  Systems}, 2001.

\bibitem{Mikolov2010}
T.~Mikolov, M.~Karafiat, L.~Burget, J.~Cernocky, and S.~Khudanpur, ``{Recurrent
  Neural Network based Language Model},'' \emph{Interspeech}, 2010.

\bibitem{Mba2016}
L.~Mba, P.~Meukam, and A.~Kemajou, ``{Application of artificial neural network
  for predicting hourly indoor air temperature and relative humidity in modern
  building in humid region},'' \emph{Energy and Buildings}, 2016.

\bibitem{Ashtiani2014}
A.~Ashtiani, P.~A. Mirzaei, and F.~Haghighat, ``{Indoor thermal condition in
  urban heat island: Comparison of the artificial neural network and regression
  methods prediction},'' \emph{Energy and Buildings}, 2014.

\bibitem{Zhu2017}
L.~Zhu and N.~Laptev, ``{Deep and Confident Prediction for Time Series at
  Uber},'' \emph{IEEE International Conference on Data Mining Workshops,
  ICDMW}, vol. 2017-November, pp. 103--110, 2017.

\bibitem{Marvuglia2014}
A.~Marvuglia, A.~Messineo, and G.~Nicolosi, ``{Coupling a neural network
  temperature predictor and a fuzzy logic controller to perform thermal comfort
  regulation in an office building},'' \emph{Building and Environment}, 2014.

\bibitem{Li2017}
Y.~Li and Y.~Gal, ``{Dropout Inference in Bayesian Neural Networks with
  Alpha-divergences},'' \emph{ICML}, 2017.

\bibitem{Gal2016}
Y.~Gal, ``{Uncertainty in Deep Learning},'' \emph{PhD Thesis}, 2016.

\bibitem{Svozil1997}
D.~Svozil, V.~Kvasni{\v{c}}ka, and J.~Posp{\'{i}}chal, ``{Introduction to
  multi-layer feed-forward neural networks},'' in \emph{Chemometrics and
  Intelligent Laboratory Systems}, 1997.

\bibitem{Dahl2012}
G.~E. Dahl, D.~Yu, L.~Deng, and A.~Acero, ``{Context-dependent pre-trained deep
  neural networks for large-vocabulary speech recognition},'' \emph{IEEE
  Transactions on Audio, Speech and Language Processing}, 2012.

\bibitem{LeCun2015}
Y.~A. LeCun, Y.~Bengio, and G.~E. Hinton, ``{Deep learning},'' \emph{Nature},
  2015.

\bibitem{GoodfellowIanBengioYoshuaCourville2016}
A.~{Goodfellow, Ian, Bengio, Yoshua, Courville}, ``{Deep Learning},'' \emph{MIT
  Press}, 2016.

\bibitem{Cao2016}
S.~Cao, W.~Lu, and Q.~Xu, ``{Deep Neural Networks for Learning Graph
  Representations},'' \emph{Aaai}, 2016.

\bibitem{Toshev2014}
A.~Toshev and C.~Szegedy, ``{DeepPose: Human pose estimation via deep neural
  networks},'' \emph{The IEEE Conference on Computer Vision and Pattern
  Recognition (CVPR)}, 2014.

\bibitem{Ciresan2012}
D.~Cireşan, U.~Meier, J.~Masci, and J.~Schmidhuber, ``{Multi-column deep
  neural network for traffic sign classification},'' \emph{Neural Networks},
  2012.

\bibitem{Huang2014}
W.~Huang, G.~Song, H.~Hong, and K.~Xie, ``{Deep architecture for traffic flow
  prediction: Deep belief networks with multitask learning},'' \emph{IEEE
  Transactions on Intelligent Transportation Systems}, 2014.

\bibitem{Romeu2013}
P.~Romeu, F.~Zamora-Mart{\'{i}}nez, P.~Botella-Rocamora, and J.~Pardo,
  ``{Time-series forecasting of indoor temperature using pre-trained deep
  neural networks},'' in \emph{Lecture Notes in Computer Science (including
  subseries Lecture Notes in Artificial Intelligence and Lecture Notes in
  Bioinformatics)}, 2013.

\bibitem{Kalogirou2000}
S.~Kalogirou, ``{Artificial neural networks for the prediction of the energy
  consumption of a passive solar building},'' \emph{Energy}, 2000.

\bibitem{Hochreiter1997}
S.~Hochreiter and J.~{Urgen Schmidhuber}, ``{LONG SHORT-TERM MEMORY},''
  \emph{Neural Computation}, 1997.

\bibitem{Chan2016}
W.~Chan, N.~Jaitly, Q.~Le, and O.~Vinyals, ``{Listen, attend and spell: A
  neural network for large vocabulary conversational speech recognition},'' in
  \emph{ICASSP, IEEE International Conference on Acoustics, Speech and Signal
  Processing - Proceedings}, 2016.

\bibitem{Cui2016}
Y.~Cui, S.~Wang, J.~Li, and Y.~Wang, ``{LSTM Neural Reordering Feature for
  Statistical Machine Translation},'' in \emph{In Proceedings of the Conference
  of the North American Chapter of the Association for Computational
  Linguistics: Human Language Technologies (NAACL-HLT)}, 2016.

\bibitem{Marino2016}
D.~L. Marino, K.~Amarasinghe, and M.~Manic, ``{Building energy load forecasting
  using Deep Neural Networks},'' in \emph{IECON 2016 - 42nd Annual Conference
  of the IEEE Industrial Electronics Society}, 2016.

\bibitem{Qiu2014}
X.~Qiu, L.~Zhang, Y.~Ren, P.~Suganthan, and G.~Amaratunga, ``{Ensemble deep
  learning for regression and time series forecasting},'' in \emph{IEEE SSCI
  2014 - 2014 IEEE Symposium Series on Computational Intelligence - CIEL 2014:
  2014 IEEE Symposium on Computational Intelligence in Ensemble Learning,
  Proceedings}, 2014.

\bibitem{Glorot2011}
X.~Glorot, A.~Bordes, and Y.~Bengio, ``{Deep sparse rectifier neural
  networks},'' \emph{AISTATS '11: Proceedings of the 14th International
  Conference on Artificial Intelligence and Statistics}, 2011.

\bibitem{Chang2007}
F.~J. Chang, Y.~M. Chiang, and L.~C. Chang, ``{Multi-step-ahead neural networks
  for flood forecasting},'' \emph{Hydrological Sciences Journal}, 2007.

\bibitem{Gal2015}
\BIBentryALTinterwordspacing
Y.~Gal and Z.~Ghahramani, ``{A Theoretically Grounded Application of Dropout in
  Recurrent Neural Networks},'' no. Nips, 2015. [Online]. Available:
  \url{http://arxiv.org/abs/1512.05287}
\BIBentrySTDinterwordspacing

\bibitem{Penciana2004}
M.~J. Penciana and R.~B. D'Agostino, ``{Overall C as a measure of
  discrimination in survival analysis: Model specific population value and
  confidence interval estimation},'' \emph{Statistics in Medicine}, 2004.

\bibitem{Sainath2015}
T.~N. Sainath, O.~Vinyals, A.~Senior, and H.~Sak, ``{Convolutional, Long
  Short-Term Memory, fully connected Deep Neural Networks},'' in \emph{ICASSP,
  IEEE International Conference on Acoustics, Speech and Signal Processing -
  Proceedings}, 2015.

\bibitem{Cho2014}
K.~Cho, B.~V. Merrienboer, D.~Bahdanau, and Y.~Bengio, ``{On the Properties of
  Neural Machine Translation : Encoder – Decoder Approaches},''
  \emph{Ssst-2014}, 2014.

\end{thebibliography}
\end{document}